\def\year{2019}\relax
\documentclass[letterpaper]{article} 
\usepackage{aaai19}  
\usepackage{times}  
\usepackage{helvet}  
\usepackage{courier}  
\usepackage{url}  
\usepackage{graphicx}  
\usepackage{amsmath}
\usepackage{amssymb}
\usepackage{makecell}
\usepackage{array}
\usepackage{multirow}
\usepackage{amsmath}
\DeclareMathOperator*{\argmax}{arg\,max}
\DeclareMathOperator{\E}{\mathbb{E}}
\usepackage{xspace}
\usepackage{booktabs}
\usepackage[font=normalsize]{subfig}
\usepackage{arydshln}
\begin{document}
%
\title{Neural Machine Translation with Adequacy-Oriented Learning}
\author{Xiang Kong \\ Carnegie Mellon University \\ {\tt xiangk@andrew.cmu.edu} \And
       Zhaopeng Tu\thanks{~~~Zhaopeng Tu is the corresponding author. Work was mainly done when Xiang Kong was interning at Tencent AI Lab.} \\ Tencent AI Lab \\ {\tt zptu@tencent.com} \AND
       Shuming Shi \\ Tencent AI Lab \\ {\tt shumingshi@tencent.com} \And
       Eduard Hovy \\ Carnegie Mellon University \\ {\tt hovy@cs.cmu.edu} \And
       Tong Zhang \\ Tencent AI Lab \\ {\tt bradymzhang@tencent.com}}

\maketitle
\begin{abstract}
Although Neural Machine Translation (NMT) models have advanced state-of-the-art performance in machine translation, they face problems like the inadequate translation. 
We attribute this to that the standard Maximum Likelihood Estimation (MLE) cannot judge the real translation quality due to its several limitations. 
In this work, we propose an adequacy-oriented learning mechanism for NMT by casting translation as a stochastic policy in Reinforcement Learning (RL), where the reward is estimated by explicitly measuring translation adequacy.
Benefiting from the sequence-level training of RL strategy and a more accurate reward designed specifically for translation, our model outperforms multiple strong baselines, including (1) standard and coverage-augmented attention models with MLE-based training, and (2) advanced reinforcement and adversarial training strategies with rewards based on both word-level BLEU and character-level \textsc{chrF3}.
Quantitative and qualitative analyses on different language pairs and NMT architectures demonstrate the effectiveness and universality of the proposed approach.

\end{abstract}

\section{Introduction}

During the past several years, rapid progress has been made in the field of Neural Machine Translation (NMT)~\cite{kalchbrenner2013recurrent,sutskever2014sequence,bahdanau2014neural,gehring2017convolutional,Wu:2016:arXiv,vaswani2017attention}. 

Although NMT models have advanced the community, they still face inadequate translation problems: one or multiple parts of the input sentence are not translated~\cite{tu2016modeling}. We attribute this problem to the lack of the mechanism to guarantee the generated translation being as sufficient as human translation. 
NMT models are generally trained in an end-to-end manner to maximize the likelihood of the output sentence.
Maximum Likelihood Estimation (MLE), however, could not judge the real quality of generated translation due to its several limitations： 
\begin{enumerate}
    \item {\em Exposure bias}~\cite{Ranzato:2016:ICLR}: The models are trained on the groundtruth data distribution, but at test time are used to generate target words based on previous model predictions, which can be erroneous;
    \item {\em Word-level loss}~\cite{Shen:2016:ACL}: Likelihood is defined at word-level, which might hardly correlate well with sequence-level evaluation metrics like BLEU. 
    \item {\em Focusing more on fluency than adequacy}~\cite{Tu:2017:AAAI}: Likelihood does not measure how well the complete source information is transformed to the target side, thus does not correlate well with translation adequacy. Adequacy metric is regularly employed to assess the translation quality in practice. 
\end{enumerate}

Some recent work partially alleviates one or two of the above problems with advanced training strategies.
For example, the first two problems are tackled by sequence level training using the REINFORCE algorithm~\cite{Ranzato:2016:ICLR,bahdanau2016actor}, minimum risk training~\cite{Shen:2016:ACL}, beam search optimization~\cite{Wiseman:2016:EMNLP} or adversarial learning~\cite{wu2017adversarial,Yang:2018:NAACL}. The last problem can be alleviated by introducing an auxiliary reconstruction-based training objective to measure translation adequacy~\cite{Tu:2017:AAAI}.

In this work, we aim to fully solve all the three problems in a unified framework. Specifically, we model the translation as a stochastic policy in Reinforcement Learning (RL) and directly perform gradient policy update.
~The RL reward is estimated on a complete sequence produced by the NMT model, which is able to correlate well with a sequence-level task-specific metric. To explicitly measure translation adequacy, we propose a novel metric called {\em Coverage Difference Ratio} (\textsc{Cdr}) which is calculated by counting how many source words are under-translated via directly comparing generated translation with human translation.
Benefiting from the sequence-level training of RL strategy and a more accurate reward designed specifically for translation, the proposed approach is able to alleviate all the aforementioned limitations of MLE-based training.

We conduct experiments on Chinese$\Rightarrow$English and German$\Leftrightarrow$English translation tasks, using both the RNN-based NMT model~\cite{bahdanau2014neural} and the recently proposed \textsc{Transformer}~\cite{vaswani2017attention}. The consistent improvements across language pairs and NMT architectures demonstrate the effectiveness and universality of the proposed approach. 
The proposed adequacy-oriented learning improves translation performance not only over a standard attention model, but also over a coverage-augmented attention model~\cite{tu2016modeling} that alleviates the inadequate translation problem at the word-level. In addition, the proposed metric -- \textsc{Cdr} score, consistently outperforms the commonly-used word-level BLEU~\cite{papineni2002bleu} and character-level \textsc{chrF3}~\cite{popovic2015chrf} scores in both the reinforcement learning and adversarial learning frameworks, indicating the superiority and necessity of an adequacy-oriented metric in training effective NMT models.

\section{Background}

Neural Machine Translation (NMT) is an end-to-end structure which could directly model the translation probability between a source sentence ${\bf x}=x_{1}, x_{2}, \dots, x_{J}$ and a target sentence ${\bf y}=y_{1}, y_{2}, \dots, y_{I}$ word by word:
\begin{equation}
    P({\bf y}|{\bf x})=\prod_{i=1}^{I}P(y_{i}|{\bf y}_{< i}, {\bf x}; {\bf \theta})
\end{equation}
where ${\bf y}_{<i}$ is the partial translation before decoding step $i$ and ${\bf \theta}$ is parameters of the NMT. The probability of generating the $i$-th word $P(y_{i}|y_{< i}, {\bf x};\theta)$ is calculated by
\begin{equation}
    P(y_{i}|{\bf y}_{< i},{\bf x};\theta)\propto \exp\left \{ f(y_{i-1}, {\bf s}_{i}, {\bf c}_{i}; {\bf \theta})\right \}
\end{equation}
where ${\bf s}_i$ is the $i$-th hidden state of the decoder and $f(\cdot)$ is a non-linear activation function of the decoder state. ${\bf c}_i$ is a distinct source representation for time $i$, calculated as a weighted sum of the source annotations: ${\bf c}_i = \sum_{j=1}^{J}{\alpha_{i,j}\cdot {\bf h}_j}$,
where ${\bf h}_j$ is the annotation of $x_j$ from a encoder, and its weight $\alpha_{i,j}$ is computed by
\begin{equation}
\alpha_{i,j} = \frac{\exp(e_{i,j})}{\sum_{j'=1}^{J} \exp(e_{i,j'})} \ \ with \ \  e_{i,j} = a({\bf s}_{i-1}, {\bf h}_j)
\label{eqn-alignment-probability} 
\end{equation}
where $a(\cdot)$ is an \emph{attention model} that scores how well $y_i$ and ${\bf h}_j$ (\emph{i.e.,}\xspace $x_j$) match. 
The encoder and decoder can be implemented as Recurrent Neural Network (RNN)~\cite{bahdanau2014neural}, Convolutional Neural Network (CNN)~\cite{gehring2017convolutional}, or Self-Attention Network (SAN)~\cite{vaswani2017attention}.

The parameters of the NMT $\theta$ are trained to maximize the likelihood of training instances $\{[x^{n},y^{n}]\}_{n=1}^{N}$:
\begin{equation}\label{eqn:mle}
    L(\theta) = \argmax_{\theta}\sum_{n=1}^{N}\log P(y^{n}|x^{n};\theta)
\end{equation}
Although likelihood is a widely-used training objective for its simpleness and effectiveness, it has several aforementioned limitations including {\em exposure bias}~\cite{Ranzato:2016:ICLR,Wiseman:2016:EMNLP}, {\em word-level estimation}~\cite{Shen:2016:ACL}, and {\em focusing more on fluency than adequacy}~\cite{Tu:2017:AAAI}.

\section{Approach}

\subsection{Intuition}


\noindent In this work, we try to solve the three problems mentioned above in a unified framework.
Our objective is three-fold:
\begin{enumerate}
    \item We solve the exposure bias problem by modeling the translation as a stochastic policy in reinforcement learning (RL) and directly performing policy gradient update.
    \item The RL reward is estimated on a complete sequence, which correlates well with either sequence-level BLEU or a more adequacy-oriented metric, as described below.
    \item We design a sequence-level metric -- {\em Coverage Difference Ratio} (\textsc{Cdr}) -- to explicitly measure translation adequacy which focuses on the commonly-cited  weaknesses of NMT models: producing fluent yet inadequate translations. We expect that the model can benefit from linguistic insights that correlate well with human intuitions.
\end{enumerate}

\begin{figure}[t]
\centering
\subfloat[\small Example of human (\textsc{Ref}) and generated (\textsc{Nmt}) translations.]{
\includegraphics[width=0.48\textwidth]{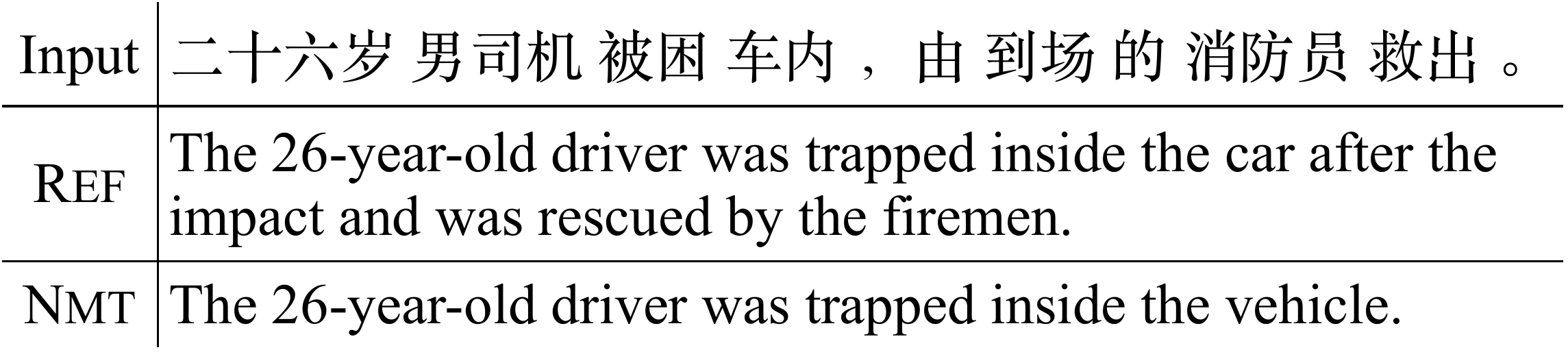}
}
\\
\subfloat[\small Examples of covered source words (i.e. shadow boxes).]{
\includegraphics[width=0.48\textwidth]{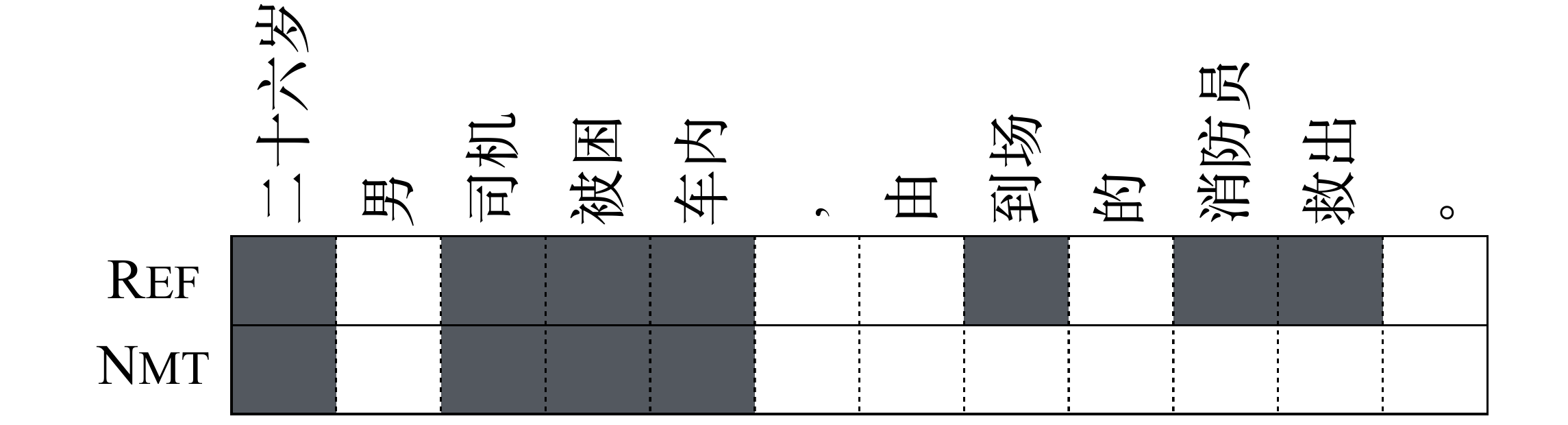}
}
\caption{An example to illustrate coverage difference ratio.}
\label{fig-example}

\end{figure}

\paragraph{Coverage Difference Ratio (\textsc{Cdr})}
We measure translation adequacy by the number of under-translated words via comparing generated translation with human translation.
We take an example to illustrate how to measure translation adequacy in terms of coverage difference ratio. Figure~\ref{fig-example}(a) shows one inadequate translation.
Following~\cite{luong2015effective,tu2016modeling}, we extract only one-to-one alignments (hard alignments) by selecting the source word with the highest alignment for each target word from the word alignments produced by NMT models.\footnote{For generated translations, we directly use the attention probability distributions from decoding procedure; for human translations, we obtain attention distributions by force decoding the target sentences with the same NMT model.} 
A source word is considered to be translated when it is covered by the hard alignments, as shown in Figure~\ref{fig-example}(b). 
Comparing source words covered by generated translation with those covered by human translation, we can find that the two sets are very different for inadequate translation. 
Specifically, the difference generally lies in the untranslated source words that cause inadequate translation problem, indicating that coverage difference ratio is a good way to measure the adequacy of generated translation.

Formally, we calculate the \textsc{Cdr} score of a given generated translation ${\bf \hat{y}}$ by
\begin{eqnarray}\label{eqn:utr}
    \textsc{Cdr} ({\bf \hat{y}} | {\bf y}, {\bf x}) &=& 1 - \frac{|{\bf C}_{ref} \setminus {\bf C}_{gen}|}{|{\bf C}_{ref}|}
\end{eqnarray}
where ${\bf C}_{ref}$ and ${\bf C}_{gen}$ is the set of source words covered by human translation and generated translation, respectively. ${\bf C}_{ref}\setminus{\bf C}_{gen}$ denotes the covered source words in ${\bf C}_{ref}$ but not in ${\bf C}_{gen}$. We use ${\bf C}_{ref}$ as the reference coverage to eliminate the effect of null-aligned source words
~which are not aligned to any target word. As seen, $\textsc{Cdr} ({\bf \hat{y}} | {\bf y}, {\bf x})$ is a number between 0 and 1, where 1 means ``completely adequate translation'' and 0 means ``completely inadequate translation''. Taking Figure~\ref{fig-example}(b) as an example, the \textsc{Cdr} score is $1-(7-4)/7=0.57$.


\subsection{\textbf{Architecture}}

\begin{figure}[t]
 \centering
 \includegraphics[width=0.4\textwidth]{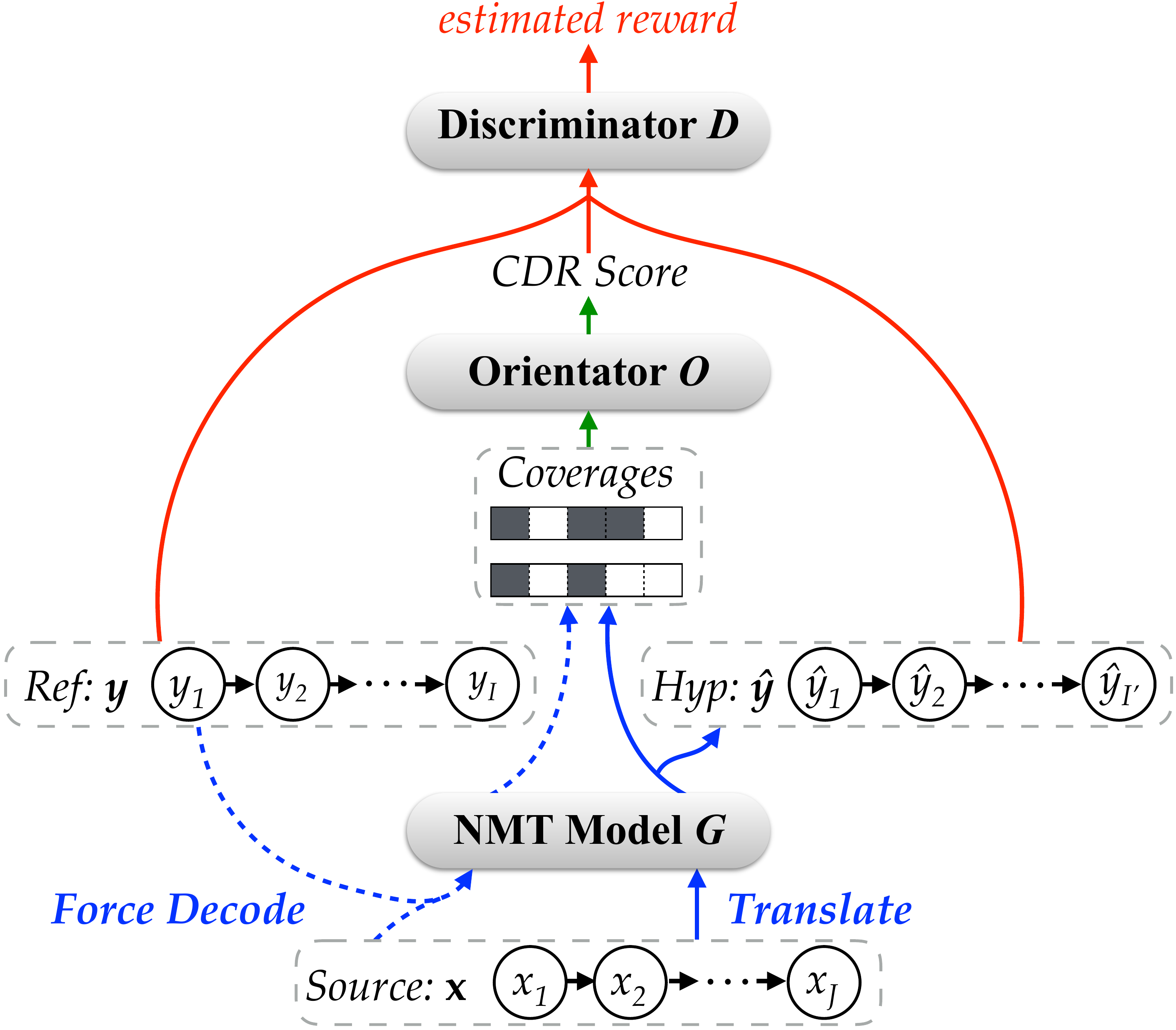}
 \caption{Architecture of adequacy-oriented NMT. The newly added orientator {\bf \em O} reads coverages of generated and human translations to generate  a \textsc{Cdr} score for each generated translation, which guides the discriminator {\bf \em D} to differentiate good generated translations from bad ones.}
 \label{figure-architecture}
 
\end{figure}

As shown in Figure~\ref{figure-architecture}, the proposed model consists of a {\em generator}, a {\em discriminator}, and an {\em orientator}.

\paragraph{Generator} The generator {\bf \em G} generates the translation ${\bf \hat{y}}$ conditioned on the input sentence ${\bf x}$. Because we need word alignments to calculate adequacy  scores in terms of \textsc{Cdr}, an attention-based NMT model is employed as the generator.

\paragraph{Orientator} The orientator {\bf \em O} reads the word alignments produced by NMT attention model when generating (or force decoding) the two translations and outputs an adequacy score for the generated translation in terms of the aforementioned \textsc{Cdr} score. Then, the orientator is used to guide the discriminator to distinguish adequate translation from inadequate ones. Accordingly, adequate translations with higher \textsc{Cdr} scores would contribute more to parameter tuning, as described in the following section.

\paragraph{Discriminator} We employ a RNN-based discriminator to differentiate generated translation from human translation, given the input sentence. The discriminator reads the input sentence ${\bf x}$ and its translation (either ${\bf y}$ or ${\bf \hat{y}}$), and use two RNNs to summarize the two sentences individually. The concatenation of the two summarized representation vectors is fed into a fully-connected neural network.

\subsection{Adequacy-Oriented Training}

In order to train the system efficiently and effectively, we employ a periodical training strategy, which is commonly used in adversarial training~\cite{goodfellow2014generative,wu2017adversarial}. Specifically, we optimize two networks with two objective functions and periodically freeze the parameters of each network during training.

\paragraph{Train Generator and Freeze Discriminator}
Following~\citeauthor{wu2017adversarial}~\shortcite{wu2017adversarial}, we use the REINFORCE algorithm~\cite{williams1992simple} to back-propagate the error signals from {\bf \em D} to {\bf \em G}, given the discretely generated ${\bf \hat{y}}$ from {\bf \em G}. The objective of the generator is to maximize the expected reward:
\begin{equation}
   L  = \E_{({\bf x}, {\bf \hat{y}}) \in G_{\theta}}[D({\bf x}, {\bf \hat{y}})] 
\end{equation}
whose gradient is
 \begin{equation}
   {\triangledown_{\theta}}
= {E_{({\bf x}, {\bf \hat{y}}) \in G_{\theta}}}[D({\bf x},{\bf \hat{y}}) {\triangledown_{{\theta}}}\log G_{\theta}({\bf \hat{y}}|{\bf x})] 
\end{equation}
The gradient is approximated by a sample from {\bf \em G} using the REINFORCE algorithm~\cite{williams1992simple}:
\begin{equation}
   {\triangledown_{\theta}} \approx {\hat{\triangledown}_{\theta}} = D({\bf x},{\bf \hat{y}}){\triangledown_{{\theta}}} \log G_{\theta}({\bf \hat{y}}|{\bf x}) 
   \label{eqn-gradient}
\end{equation}
where ${\triangledown_{{\theta}}} \log G({\bf \hat{y}}|{\bf x})$ is the standard NMT gradient which is calculated by the maximum likelihood estimation.
Therefore, the final update function for the generator is:
\begin{equation}
   {\theta}={\theta}- \eta {\hat{\triangledown}_{\theta}} 
\end{equation}
where the $\eta$ is the learning rate.
Based on the update function, 
when the $D({\bf x},{\bf \hat{y}})$ is large (\emph{i.e.,}\xspace ideally, the generated translation ${\bf \hat{y}}$ has a high adequacy score) 
, the larger reward the NMT model will get, and thus parameters are updated more based on the adequate training instance $({\bf x},{\bf \hat{y}})$.

\paragraph{Train Discriminator {\em Oriented by Adequacy} and Freeze Generator}
Ideally, a good translation ${\bf \hat{y}}$ should be assigned a high adequacy score $D({\bf x},{\bf \hat{y}})$ and thus contribute more to updating the generator.
Therefore, we expect the discriminator to not only differentiate generated translations from human translations but also distinguish bad generated translations from good ones. 
Therefore, a new objective of discriminator is to assign a precise score for each generated translation, which is consistent with their adequacy score:
\begin{equation}
  \min_{D}|\textsc{Cdr}({\bf \hat{y}}|{\bf x}, {\bf y}) - D({\bf x}, {\bf \hat{y}})|^{2}
\end{equation}
where $\textsc{Cdr}({\bf \hat{y}}|{\bf x}, {\bf y})$ is the coverage difference ratio of ${\bf \hat{y}}$. As seen, a well trained discriminator would assign a distinct score to each generated translation, which can better measure its adequacy.
\section{Related Work}


This work is related to modeling translation as policy gradient and adequacy modeling. For the former, we take minimum risk training, reinforcement learning and adversarial learning as representative strategies.

\paragraph{Minimum Risk Training}
In response to the exposure bias and word-level loss problems of MLE training,~\citeauthor{Shen:2016:ACL}~\shortcite{Shen:2016:ACL} minimize the expected loss in terms of evaluation metrics on the training data. 
Our simplified model is analogous to their MRT model, if we directly use \textsc{Cdr} as the reward to update parameters:
\begin{equation}
   {\hat{\triangledown}_{\theta}} = \textsc{Cdr}({\bf \hat{y}}|{\bf x},{\bf {y}})){\triangledown_{{\theta}}} \log G_{\theta}({\bf \hat{y}}|{\bf x}) 
\end{equation}
The simplified model differs in that (1) we use adequacy-oriented metric (\emph{i.e.,}\xspace \textsc{Cdr}) while they use sequence-level BLEU, and (2) we only need to sample one candidate to calculate reinforcement reward while they generate multiple samples to calculate the expected risk.
In addition, our discriminator gives a smoother and dynamically-updated objective compared with directly using the adequacy-oriented metric, because the latter is highly sensitive to the slight coverage difference~\cite{koehn2017six}.

\paragraph{Reinforcement Learning}
Recent work shows that maximum likelihood training could be sub-optimal due to the different conditions between training and test modes~\cite{bengio2015scheduled,Ranzato:2016:ICLR}. In order to address the exposure bias and the loss which does not operate at the sequence level, \citeauthor{Ranzato:2016:ICLR}~\shortcite{Ranzato:2016:ICLR} employ the REINFORCE algorithm~\cite{williams1992simple} to decide whether or not tokens from a sampled prediction could contribute to a high task-specific score (\emph{e.g.,}\xspace BLEU). \citeauthor{bahdanau2016actor}~\shortcite{bahdanau2016actor} use the actor-critic method from reinforcement learning to directly optimize a task-specific score. 


\paragraph{Adversarial Learning}
Recently, adversarial learning~\cite{goodfellow2014generative} has been successfully applied to neural machine translation~\cite{wu2017adversarial,Yang:2018:NAACL,Cheng:2018:ACL}.
In the adversarial framework, NMT models generally serve as the generator which defines the policy to generate the target sentence {\bf y} given the source sentence {\bf x}.
A discriminator tries to distinguish the translation result ${\bf \hat{y}} = G({\bf x})$ from the human-generated one ${\bf y}$, given the source sentence ${\bf x}$.


If we remove the orientator {\bf \em O}, our model is roll-backed to the adversarial NMT, and the training objective of the discriminator {\bf \em D} is rewritten as
\begin{eqnarray}
    \max_{D} \{\log D({\bf x}, {\bf y}) + \log (1-D({\bf x}, {\bf \hat{y}})) \} 
\end{eqnarray}
The goal of the discriminator is try to maximize the likelihood of human translation $D({\bf x}, {\bf y})$ to 1 and minimize that of generated translation $D({\bf x}, {\bf \hat{y}})$ to 0.

As seen, the discriminator uses a binary classification by uniformly treating all generated translations as negative examples (\emph{i.e.,}\xspace labeling ``0'') and all human translations as positive examples (\emph{i.e.,}\xspace labeling ``1''), regardless of the quality of the generated translations. However, intuitively, high-quality translations and low-quality translations should be treated differently by the discriminator, otherwise, inaccurate reward signals would be propagated back to the generator. In our proposed architecture, this problem can be alleviated by replacing the simple binary outputs with the more informative adequacy-oriented metric \textsc{Cdr}, which is calculated by directly comparing generated and human translations.

\paragraph{Adequacy Modeling}
Inadequate translation problem is a commonly-cited weakness of NMT models~\cite{tu2016modeling}. A number of recent efforts have explored ways to alleviate this problem. For example,~\citeauthor{tu2016modeling}~\shortcite{tu2016modeling} and~\citeauthor{Mi:2016:EMNLP}~\shortcite{Mi:2016:EMNLP} employ coverage vector as a lexical-level indicator to indicate whether a source word is translated or not.~\citeauthor{Zheng:2018:TACL}~\shortcite{Zheng:2018:TACL} and ~\citeauthor{Meng:2018:IJCAI}~\shortcite{Meng:2018:IJCAI} move one step further and directly model translated and untranslated source contents by operating on the attention context vector.~\citeauthor{He:2017:NIPS}~\shortcite{He:2017:NIPS} use a prediction network to estimate the future cost of translating the uncovered source words.
Our approach is complementary to theirs since they model the adequacy learning at the word-level inside the generator (i.e., NMT models), while we model it at the sequence-level outside the generator.
We take the representative coverage mechanism~\cite{tu2016modeling} as another stronger baseline model for its simplicity and efficiency, and experimental results show that our model can further improve performance.

In the context of adequacy-oriented training,~\citeauthor{Tu:2017:AAAI}~\shortcite{Tu:2017:AAAI} introduce an auxiliary objective to measure the adequacy of translation candidates, which is calculated by reconstructing generated translations back to the original inputs. Benefiting from the flexible framework of reinforcement training, we are able to directly compare generated translations with human translations and define a more straightforward metric, \emph{i.e.,}\xspace \textsc{Cdr} to measure adequacy of generated sentences.

\begin{table*}[t]
  \centering

  \begin{tabular}{c|l||r r||cccc|l}
    \bf \#  &   \bf Model &\bf \# Para. & \bf Speed &  \bf  MT03 & \bf MT04 & \bf MT05 & \bf MT06 &  \bf All \\
    \hline
    1   &   \textsc{RNNSearch} & 86.7M & 1.4K  & 36.00 & 38.26 &35.88 & 35.98 & 36.76   \\ 
    2 & ~~~ + MRT$_{\textsc{Bleu}}$ & +0M & 0.3K & 37.32 & 39.41 & 36.78 & 37.22 & 37.92$^{\ddag}$  \\
    \hline
    3   &   ~~~ + {\bf \em D}$_{\textsc{Cnn}}$	& +0.23M & 1.0K &  37.11 & 38.84 & 35.97&37.36 &37.54$^{\ddag}$ \\ 
    4   &   ~~~ + {\bf \em D}$_{\textsc{Rnn}}$	& +0.17M & 1.2K &36.96 & 38.92 & 36.13 & 37.54 & 37.59$^{\ddag}$  \\ 
    \hline
    5   &   ~~~ + {\bf \em O}$_{\textsc{Bleu}}$ & +0M & 1.1K& 37.02&38.49&36.62&36.21& 37.44$^{\dag}$ \\
    6   &   ~~~ + {\bf \em O}$_{\textsc{chrF3}}$ & +0M & 1.1K& 37.91&39.80&36.57&35.95& 37.75$^{\dag}$ \\
    7   &   ~~~ + {\bf \em O}$_{\textsc{Cdr}}$ & +0M & 1.0K	& 37.42&39.52&36.86&37.34& 38.02$^{\ddag}$\\ 
    \hline
    8   &   ~~~ + {\bf \em D}$_{\textsc{Rnn}}$ + {\bf \em O}$_{\textsc{Cdr}}$ & +0.17M & 0.8K & 37.61&40.05&37.58&36.87 &  38.42$^{\ddag}$\\
    \hline \hline
    9   &   \textsc{RNNSearch-Coverage} &  +1.03M   & 1.2K  &   38.04   & 41.09 &   38.73 &   36.52   & 39.13  \\
    10   &   ~~~ + {\bf \em D}$_{\textsc{Rnn}}$ + {\bf \em O}$_{\textsc{Cdr}}$  &  +1.20M   &  0.7K & 38.62   &   41.98   &   39.39   &   37.42   & 39.81$^{\dag}$\\
    
  \end{tabular}
  \caption{Evaluation of translation performance on Zh$\Rightarrow$En translation. ``{\bf \em D}'' denotes discriminator and ``{\bf \em O}'' denotes orientator. ``{MRT}'' indicates minimum risk training~\cite{Shen:2016:ACL}, and ``{\bf \em D}$_{\textsc{Cnn}}$'' indicates adversarial training with a CNN-based discriminator~\cite{wu2017adversarial}.
  ``\# Para.'' denotes the number of parameters, and ``Speed'' denotes the training speed (words/second). ``$\dag$'' and ``$\ddag$'' indicate statistically significant difference ($p < 0.05$ and $p < 0.01$ respectively) from the corresponding baseline.} 
\label{tab:res} 
\end{table*}

\section{Experiments}

\subsection{Setup}

We conduct experiments on the widely-used Chinese (Zh) $\Rightarrow$English (En) and German (De) $\Leftrightarrow$English (En) translation tasks. For Zh$\Rightarrow$En translation, the training corpus contains 1.25M sentence pairs extracted from LDC corpora. 
NIST 2002 (MT02) dataset is the validation set and the test data consists of NIST 2003 (MT03), NIST2004 (MT04), NIST 2005 (MT05) and NIST 2006(MT06). 
For De$\Leftrightarrow$En translation, to compare with the results reported by previous work~\cite{Shen:2016:ACL,bahdanau2016actor,wu2017adversarial,vaswani2017attention}, we use both the IWSLT 2014 and WMT 2014 data. The former contains 153K sentence pairs and the latter consists of 4.56M sentence pairs.
The 4-gram NIST BLEU score \cite{papineni2002bleu} is used as the evaluation metric and sign-test~\cite{collins2005clause} is employed to test statistical significance. 


For training all neural models, we set the vocabulary size to 30K for Zh$\Rightarrow$En, for IWSLT 2014 De$\Rightarrow$En, we follow the preprocessing procedure as used in ~\citeauthor{Ranzato:2016:ICLR}~\shortcite{Ranzato:2016:ICLR} and for WMT 2014 En$\Rightarrow$De, preprocessing method described in~\citeauthor{vaswani2017attention}~\shortcite{vaswani2017attention} is borrowed. We pre-train the discriminator on translation samples produced by the pre-trained generator. After that, the discriminator and the generator are trained together, and the generator is updated by the REINFORCE algorithm mentioned above. We also follow the training tips  mentioned in ~\citeauthor{Shen:2016:ACL}~\shortcite{Shen:2016:ACL} and ~\citeauthor{wu2017adversarial}~\shortcite{wu2017adversarial}. The hyper-parameter $\alpha$ which could control the sharpness of the generator distribution in our system is $1e\scalebox{0.75}[1.0]{-}4$, which could also be regarded as a baseline to reduce the variance of the REINFORCE algorithm. We also randomly choose 50\% minibatches trained with our objective function and the other with the MLE principle.
In MRT training strategy~\cite{Shen:2016:ACL}, the sample size is 25, the hyper-parameter $\alpha$ is $5e\scalebox{0.75}[1.0]{-}3$ and the loss function is negative smoothed sentence-level BLEU.

We validate our models on two representative model architectures, namely \textsc{RNNSearch} and \textsc{Transformer}.
For the \textsc{RNNSearch} model, mini-batch size is 80, 
the word-embedding dimension is 620, and the hidden layer size is 1000. We use a neural coverage model for \textsc{RNNSearch-Coverage} and the dimensionality of coverage vector is 100. The baseline models are trained for 15 epochs, 
which are used as the initial generator in the proposed framework.
For the \textsc{Transformer} model, we implement our proposed approach on top of an open source toolkit THMUT~\cite{zhang2017thumt}. Configurations in~\citeauthor{vaswani2017attention}~\shortcite{vaswani2017attention} are used to train the baseline models.

\begin{table*}[t]
\centering
\begin{tabular}{l|l|l}
{\bf System} &  {\bf Model}  &   {\bf De$\Rightarrow$En}     \\
\hline
\hline
\multicolumn{3}{c}{\em Existing end-to-end NMT systems} \\
\hline
\cite{Ranzato:2016:ICLR}    &   CNN encoder + Sequence level objective  &   20.73 \\
\cite{bahdanau2016actor}    &   CNN encoder + Actor-critic &   22.45    \\
\cite{Wiseman:2016:EMNLP}   &   \textsc{RNNSearch} + Beam search optimization   &   25.48   \\
\cite{wu2017adversarial} & \textsc{RNNSearch} + Adversarial objective &   26.98  \\
\hline
\hline
\multicolumn{3}{c}{\em Our end-to-end NMT systems} \\
\hline
\multirow{5}{*}{\em this work}  &   \textsc{RNNSearch}    &   26.51 \\
                                &    ~~~~ + MRT$_{\textsc{Bleu}}$ & 27.29$^{\dag}$  \\
                                &   ~~~~~+ {\bf \em D}$_{\textsc{Cnn}}$ &  27.24$^{\dag}$  \\
                                &   ~~~~~+ {\bf \em O}$_{\textsc{Cdr}}$ & 27.31$^{\dag}$   \\
                                &   ~~~~~+ {\bf \em D}$_{\textsc{Rnn}}$ + {\bf \em O}$_{\textsc{Cdr}}$ &   27.79$^{\ddag}$  \\
\end{tabular}
\caption{Comparing with previous works of applying reinforcement learning for NMT on IWSLT 2014 De$\Rightarrow$En translation task. ``$\dag$'' and ``$\ddag$'' indicate statistically significant difference ($p < 0.05$ and $p < 0.01$ respectively)  from the  \textsc{RNNSearch} model. }
\label{tab:res-iwslt-en-de} 
\end{table*}

\begin{table}[t]
\centering
\small
\begin{tabular}{l|l}
\hline
{\bf Model}  &   {\bf BLEU}    \\
\hline
GNMT + RL \cite{Wu:2016:arXiv}  &  26.30 \\
ConvS2S \cite{gehring2017convolutional}  & 26.43\\
Transformer (Base) \cite{vaswani2017attention}  & 27.3 \\
Transformer (Big) \cite{vaswani2017attention} & 28.4 \\
\hline
  \textsc{Transformer-Base}    &   27.30 \\
~~~~~+ {\bf \em O}$_{\textsc{Cdr}}$ & 27.80 \\
 ~~~~~+ {\bf \em D}$_{\textsc{Rnn}}$ + {\bf \em O}$_{\textsc{Cdr}}$ & 28.01$^{\dag}$  \\\cline{1-2}
\textsc{Transformer-Big}    &   28.35 \\
~~~~~+ {\bf \em O}$_{\textsc{Cdr}}$ & 28.63 \\
~~~~~+ {\bf \em D}$_{\textsc{Rnn}}$ + {\bf \em O}$_{\textsc{Cdr}}$ & 28.99$^{\dag}$  \\
\hline
\end{tabular}
\caption{BLEU scores on WMT 2014 En$\Rightarrow$De testset using the state-of-the-art \textsc{Transformer} model. 
`$\dag$'' indicates statistically significant difference ($p < 0.05$)  from the corresponding \textsc{Transformer} baseline model.}
\label{tab:res-wmt-att-en-de} 

\end{table}

\subsection{Chinese-English Translation Task}

Table~\ref{tab:res} lists the results of various translation models on Zh$\Rightarrow$En corpus. As seen, all advanced systems significantly outperform the baseline system (\emph{i.e.,}\xspace \textsc{RNNSearch}), although there are still considerable differences among different variants.

\paragraph{\bf Architectures of Discriminator} (Rows 3-4)
We evaluate two architectures for the discriminator. The CNN-based discriminator is composed of two convolution layers with $3\times3$ window, two max-pooling layers with $2\times2$ window and one softmax layer. The feature map size is 10 and the feed-forward hidden size is 20. The RNN-based discriminator consists of two two-layer RNN encoders with 32 LSTM units and a fully-connected neural network with 32 units.
We find that the RNN discriminator achieves similar performance with its CNN counterpart (37.59 vs. 37.54), while has a faster training speed (1.2K vs. 1.0K words/second). 
The main reason is that the CNN-based discriminator requires high computation and space cost to utilize multiple layers with convolution and pooling from a large input matrix. 

\paragraph{\bf Adequacy Metrics for Orientator} (Rows 5-7)
As aforementioned, the \textsc{Cdr} score can be directly used as a reward to update the parameters, which is in analogy to the MRT~\cite{Shen:2016:ACL} except that we use 1-best sample while they use n-best samples. For comparison, we also used the word-level BLEU score (Row 5) and character-level \textsc{chrF3} score~\cite{popovic2015chrf} (Row 6) as the rewards.

As seen, this strategy consistently improves translation performance, without introducing any new parameters. 
The extra computation cost is mainly from generating translation sentence and force decoding the human translation with the NMT model.
We find that \textsc{Cdr} not only outperforms its 1-best counterpart ``{\bf \em O}$_{\textsc{Bleu}}$'' and ``{\bf \em O}$_{\textsc{chrF3}}$'', but also surpasses ``MRT$_{\textsc{Bleu}}$'' using 25 samples. 
We attribute this to the fact that \textsc{Cdr} can better estimate the adequacy of the translation, which is the key problem of NMT models, and go beyond the  the simple low-level n-gram matching measured by BLEU and \textsc{chrF3}.

\paragraph{\bf Combining Them Together} (Row 8)
By combining advantages of both reinforcement learning and adequacy-oriented objective, our model achieves the best performance, which is 1.66 BLEU points better than the baseline ``\textsc{RNNSearch}'',  up to 0.98 BLEU points better than using single component and significantly improve the performance of ``MRT$_{\textsc{Bleu}}$'' model.
One more observation can be made. ``+D+O'' outperforms its ``+O'' counterpart (\emph{e.g.,}\xspace 8 vs. 7), which confirms our claim that the discriminator gives a smoother and dynamically-updated score than directly using the calculated one.

\paragraph{\bf Working with Coverage Model} (Rows 11-12)
\citeauthor{tu2016modeling}~\shortcite{tu2016modeling} propose a coverage model to indicate whether a source word is translated or not, which alleviates the inadequate translation problem of NMT models. We argue that our model is complementary to theirs, because we model the adequacy learning outside the generator by using an additional adequacy-oriented discriminator, while they model it inside the generator.
Experimental results validate our hypothesis: the proposed approach further improves performance by 0.58 BLEU points over the coverage-augmented model \textsc{RNNSearch-Coverage}.

\subsection{English-German Translation Tasks}

To compare with previous work of applying reinforcement learning for NMT~\cite{Ranzato:2016:ICLR,bahdanau2016actor,Wiseman:2016:EMNLP,wu2017adversarial}, we first conduct experiments on IWSLT 2014 De$\Rightarrow$En translation task. As listed in Table~\ref{tab:res-iwslt-en-de}, we reproduce the results of adversarial training reported by~\citeauthor{wu2017adversarial}~\shortcite{wu2017adversarial} (27.24 vs. 26.98). Furthermore, the proposed approach consistently outperforms previous works, demonstrating the effectiveness of our models.

We also evaluate our model on the recently proposed \textsc{Transformer} model~\cite{vaswani2017attention} on WMT 2014 En$\Rightarrow$De corpus. As shown in Table~\ref{tab:res-wmt-att-en-de}, our models significantly improve performances in all cases.
Combining with previous results, our model consistently improve translation performance across various language pairs and NMT architectures, demonstrating the effectiveness and universality of the proposed approach.

\subsection{Analysis}

To better understand our models, we conduct extensive analyses on the Zh$\Rightarrow$En translation task. 

\begin{table}[t]
\centering
\begin{tabular}{l|cr|cr} 
  \bf Model     &  \bf \textsc{Man}   &   \bf $\bigtriangleup$ & \bf \textsc{Cdr}    &   \bf $\bigtriangleup$ \\
   \hline
  \textsc{\small RNNSearch}         &   3.31 $_{\pm 0.70}$    &   --    & 0.68  &   --   \\
  ~~~+ {\bf \em D}                  &   3.57 $_{\pm 0.61}$    &   7.9\% & 0.71  &   4.4\% \\
  ~~~+ {\bf \em O}                  &   3.69 $_{\pm 0.48}$    &   11.5\%    & 0.75  &   10.3\% \\
  ~~~+ {\bf \em D} + {\bf \em O}    &   3.79 $_{\pm 0.47}$   &   14.5\%     & 0.80  &   17.6\%  \\
\end{tabular}
 \caption{Adequacy scores on randomly selected 100 sentences on Zh$\Rightarrow$En task, which are measured by \textsc{Cdr} and human evaluation (``\textsc{Man}'').}

   \label{tab:subject-eval}
\end{table}

\paragraph{Adequacy Evaluation}
To better evaluate the adequacy, we randomly choose 100 sentences from the test set, and ask two human evaluators to judge the quality of generated translations. Five scales have been set up, i.e., $\{1,2,3,4,5\}$, where ``$1$'' means that it is irrelevant between the source sentence and the translation sentence, and ``$5$'' means that from semantic and syntactic aspect, the translation sentence and the source sentence is completely equivalent.

Table~\ref{tab:subject-eval} lists the results of human evaluation and the proposed \textsc{Cdr} score. First, our models consistently improve the translation adequacy under both human evaluation and the \textsc{Cdr} score, indicating that the proposed approaches indeed alleviate the inadequate translation problem. Second, the relative improvement on \textsc{Cdr} is consistent with that on subjective evaluation. The Pearson Correlation Coefficient between \textsc{Cdr} and manual evaluation score is 0.64, indicating that the proposed \textsc{Cdr} is a reasonable metric to measure translation adequacy.

\paragraph{Length Analysis}

\begin{figure}[t]
\centering
    \includegraphics[width=0.30\textwidth]{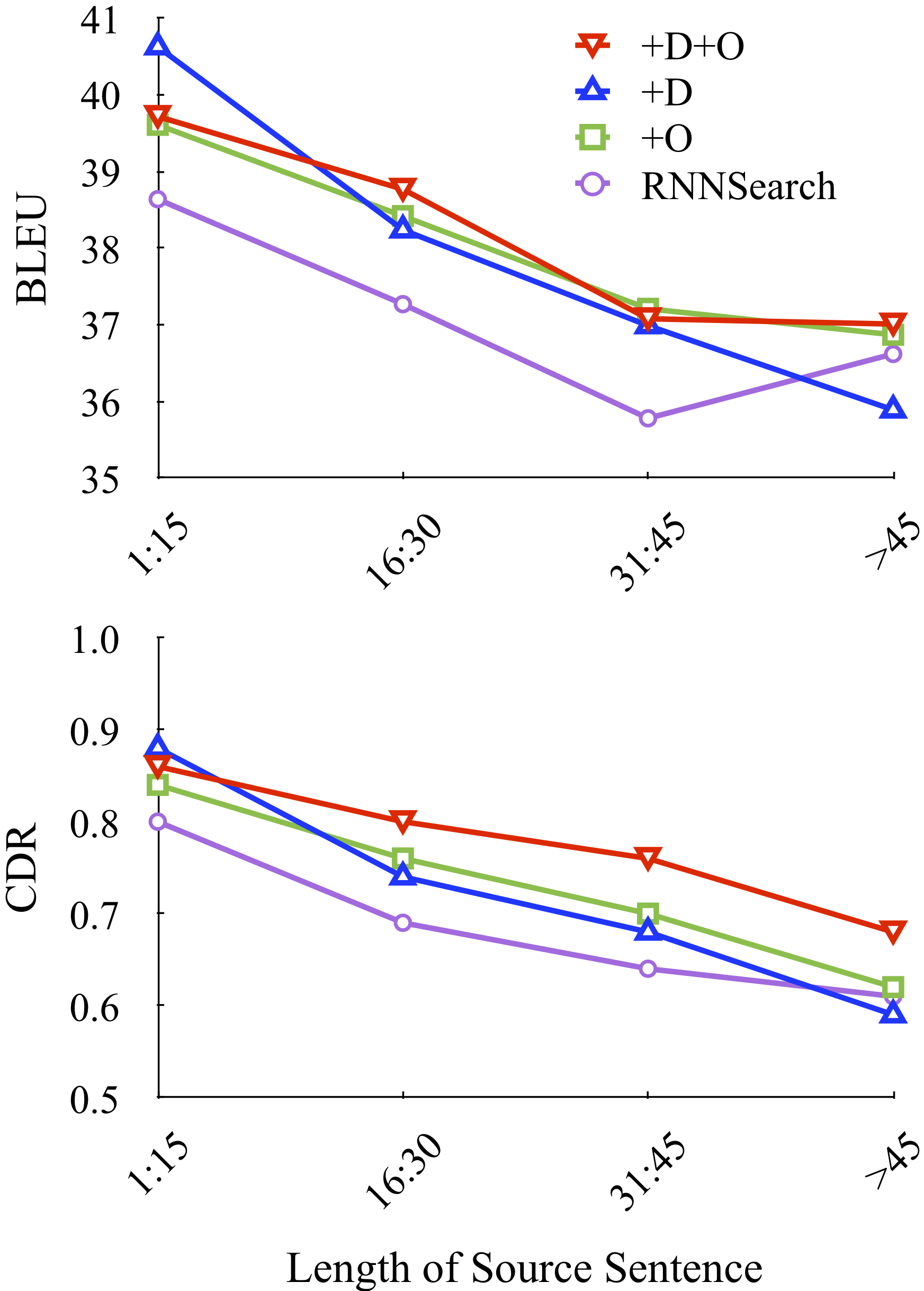}
    \caption{BLEU and \textsc{Cdr} scores of the translations with respect to the input lengths.}
    \label{figure-sentence-len}
\end{figure}

We group sentences of similar lengths and compute both the BLEU score and \textsc{Cdr} score for each group, as shown in Figure~\ref{figure-sentence-len}. 
The four length spans contain 1386, 2284, 1285, and 498 sentences, respectively.
From the perspective of the BLEU score, the proposed model (\emph{i.e.,}\xspace ``+D+O'') outperforms \textsc{RNNSearch} in all length segments. 
In contrast, using discriminator only (\emph{i.e.,}\xspace ``+D'') outperforms \textsc{RNNSearch} in most cases, except long sentences (\emph{i.e.,}\xspace $>45$). One possible reason is that it is difficult for the discriminator to differentiate generated translations from human translations for long source sentences, thus the generator cannot learn well about these instances due to the ``mistaken'' rewards from the discriminator. 
Accordingly, using the \textsc{Cdr} score (\emph{i.e.,}\xspace ``+O'') alleviates this problem by providing a sequence-level score, which better estimates the adequacy of the translations.
The final model combines the advantages of both a smoother and dynamically-updated objective from the discriminator (``+D''), and a more accurate objective specifically designed for the translation task from the orientator (``+O''). 

The \textsc{Cdr} scores for all models degrade when the length of source sentence increases. This is mainly due to that inadequate translation problem is more serious on longer sentences for NMT models~\cite{tu2016modeling}. The adversarial model (\emph{i.e.,}\xspace ``+{\bf \em D}'') improves \textsc{Cdr} scores while the improvement degrades faster with the increase of sentence length. However, our proposed approach consistently improves \textsc{Cdr} performance in all length segments.

\paragraph{Effect of the Discriminator}
\citeauthor{koehn2017six}~\shortcite{koehn2017six} point out that the attention model does not always correspond to word alignment and may considerably diverge. Accordingly, the attention matrix-based \textsc{Cdr} score  may not always correctly reflect the adequacy of generation sentences. However, our discriminator is able to give a smoother and dynamically-updated objective, and thus could provide more accurate adequacy scores of generation sentences. From the above quantitative and qualitative results, the discriminator indeed leads to better performance (\emph{i.e.,}\xspace ``+D+O'' vs. ``+O'').

\begin{figure}[t]
    \centering
    \includegraphics[width=0.4\textwidth]{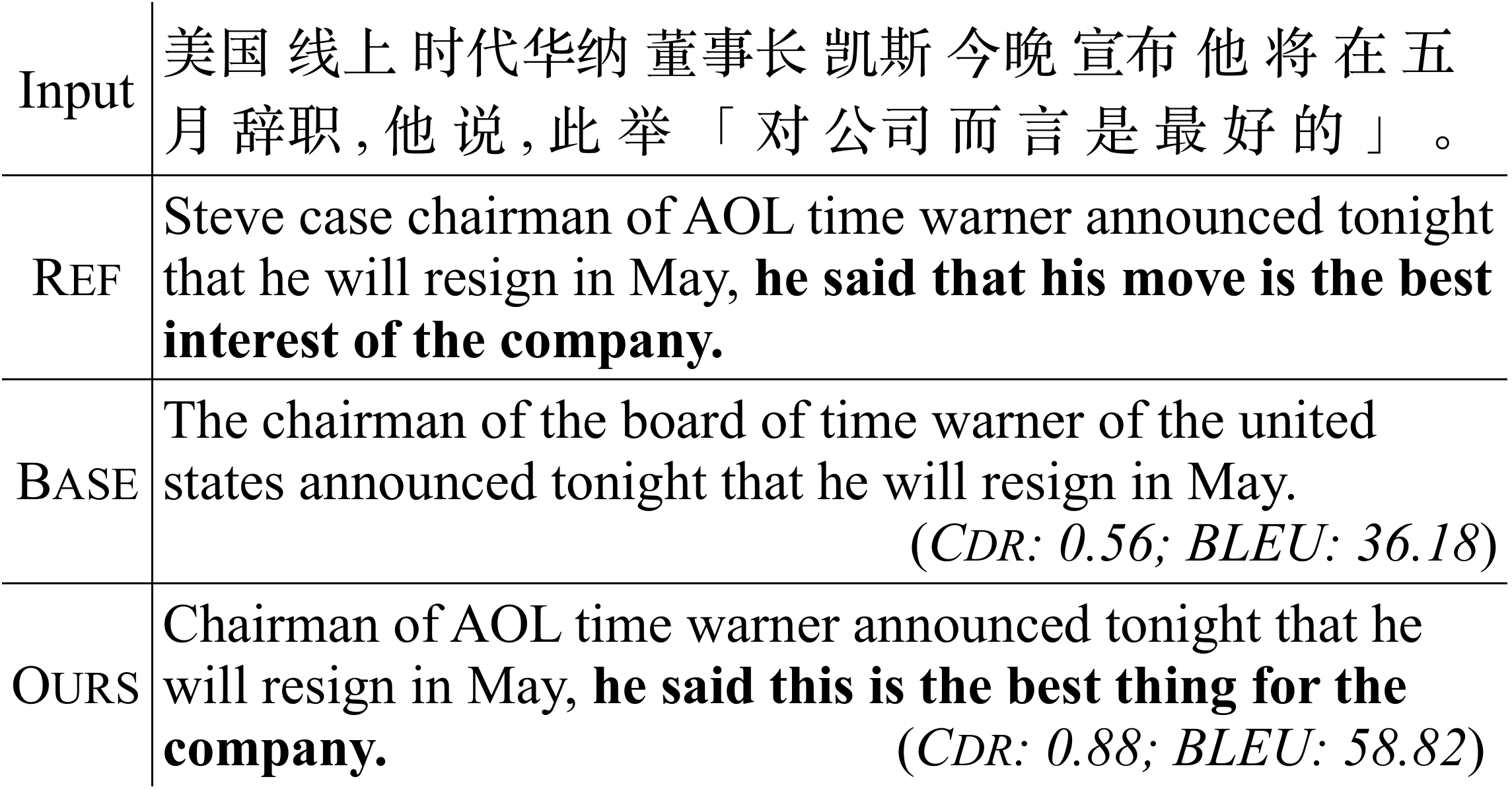}
    \caption{Example translations on Zh$\Rightarrow$En test set.}
    \label{tab:case}
\end{figure}

\paragraph{Case Study}
To better understand the advantage of our proposed model, we show a translation case in Figure~\ref{tab:case}. Specially, we provide a Zh$\Rightarrow$En example with two translation results from the RNNSearch and Adequacy-NMT models respectively, as well as the corresponding \textsc{Cdr} and BLEU scores. We emphasize on their different parts with bold fonts which lead to different translation quality.
As seen, the latter part of the source sentence is not translated by the RNNSearch model while our proposed model correct this mistake. Accordingly, our model improves both \textsc{Cdr} and BLEU scores.

\section{Conclusion}

In this work, we propose a novel learning approach for RL-based NMT models, which integrates into the policy gradient with an adequacy-oriented reward designed specifically for translation. 
The proposed approach combines the advantages of both sequence-level training of reinforcement learning, as well as a more accurately estimated reward by considering the translation adequacy in terms of coverage difference ratio (\textsc{Cdr}).
Experimental results on different language pairs show that our proposed approach not only significantly outperforms standard NMT models, but also further improves performance over those using the policy gradient and the adequacy-oriented reward individually. In addition, the proposed approach is also complementary to the coverage models~\cite{tu2016modeling}, because the two models aim to alleviate the inadequate translation problem from two different perspectives (\emph{i.e.,}\xspace sequence-level vs. word-level).

Future directions include validating our approach on other architectures such as CNN-based NMT models~\cite{gehring2017convolutional} and improved \textsc{Transformer} models~\cite{shaw2018self,Shen:2018:AAAI}, as well as combining with other advanced techniques in reinforcement learning and adversarial learning~\cite{bahdanau2016actor,Yu:2017:AAAI,Yang:2018:NAACL}.

\bibliography{mybib}

\begin{thebibliography}{}

\bibitem[\protect\citeauthoryear{Bahdanau \bgroup et al\mbox.\egroup
  }{2017}]{bahdanau2016actor}
Bahdanau, D.; Brakel, P.; Xu, K.; Goyal, A.; Lowe, R.; Pineau, J.; Courville,
  A.; and Bengio, Y.
\newblock 2017.
\newblock An actor-critic algorithm for sequence prediction.
\newblock In {\em ICLR}.

\bibitem[\protect\citeauthoryear{Bahdanau, Cho, and
  Bengio}{2015}]{bahdanau2014neural}
Bahdanau, D.; Cho, K.; and Bengio, Y.
\newblock 2015.
\newblock Neural machine translation by jointly learning to align and
  translate.
\newblock In {\em ICLR}.

\bibitem[\protect\citeauthoryear{Bengio \bgroup et al\mbox.\egroup
  }{2015}]{bengio2015scheduled}
Bengio, S.; Vinyals, O.; Jaitly, N.; and Shazeer, N.
\newblock 2015.
\newblock Scheduled sampling for sequence prediction with recurrent neural
  networks.
\newblock In {\em NIPS}.

\bibitem[\protect\citeauthoryear{Cheng \bgroup et al\mbox.\egroup
  }{2018}]{Cheng:2018:ACL}
Cheng, Y.; Tu, Z.; Meng, F.; Zhai, J.; and Liu, Y.
\newblock 2018.
\newblock Towards robust neural machine translation.
\newblock In {\em ACL}.

\bibitem[\protect\citeauthoryear{Collins, Koehn, and
  Ku{\v{c}}erov{\'a}}{2005}]{collins2005clause}
Collins, M.; Koehn, P.; and Ku{\v{c}}erov{\'a}, I.
\newblock 2005.
\newblock Clause restructuring for statistical machine translation.
\newblock In {\em ACL}.

\bibitem[\protect\citeauthoryear{Gehring \bgroup et al\mbox.\egroup
  }{2017}]{gehring2017convolutional}
Gehring, J.; Auli, M.; Grangier, D.; Yarats, D.; and Dauphin, Y.~N.
\newblock 2017.
\newblock Convolutional sequence to sequence learning.
\newblock In {\em ICML}.

\bibitem[\protect\citeauthoryear{Goodfellow \bgroup et al\mbox.\egroup
  }{2014}]{goodfellow2014generative}
Goodfellow, I.; Pouget-Abadie, J.; Mirza, M.; Xu, B.; Warde-Farley, D.; Ozair,
  S.; Courville, A.; and Bengio, Y.
\newblock 2014.
\newblock Generative adversarial nets.
\newblock In {\em NIPS}.

\bibitem[\protect\citeauthoryear{He \bgroup et al\mbox.\egroup
  }{2017}]{He:2017:NIPS}
He, D.; Lu, H.; Xia, Y.; Qin, T.; Wang, L.; and Liu, T.
\newblock 2017.
\newblock Decoding with value networks for neural machine translation.
\newblock In {\em NIPS}.

\bibitem[\protect\citeauthoryear{Kalchbrenner and
  Blunsom}{2013}]{kalchbrenner2013recurrent}
Kalchbrenner, N., and Blunsom, P.
\newblock 2013.
\newblock Recurrent continuous translation models.
\newblock In {\em EMNLP}.

\bibitem[\protect\citeauthoryear{Koehn and Knowles}{2017}]{koehn2017six}
Koehn, P., and Knowles, R.
\newblock 2017.
\newblock Six challenges for neural machine translation.
\newblock In {\em Proceedings of the First Workshop on Neural Machine
  Translation},  28--39.

\bibitem[\protect\citeauthoryear{Luong, Pham, and
  Manning}{2015}]{luong2015effective}
Luong, M.-T.; Pham, H.; and Manning, C.~D.
\newblock 2015.
\newblock Effective approaches to attention-based neural machine translation.
\newblock In {\em EMNLP}.

\bibitem[\protect\citeauthoryear{Meng \bgroup et al\mbox.\egroup
  }{2018}]{Meng:2018:IJCAI}
Meng, F.; Tu, Z.; Cheng, Y.; Wu, H.; Zhai, J.; Yang, Y.; and Wang, D.
\newblock 2018.
\newblock Neural machine translation with key-value memory-augmented attention.
\newblock In {\em IJCAI}.

\bibitem[\protect\citeauthoryear{Mi \bgroup et al\mbox.\egroup
  }{2016}]{Mi:2016:EMNLP}
Mi, H.; Sankaran, B.; Wang, Z.; and Ittycheriah, A.
\newblock 2016.
\newblock Coverage embedding models for neural machine translation.
\newblock In {\em EMNLP}.

\bibitem[\protect\citeauthoryear{Papineni \bgroup et al\mbox.\egroup
  }{2002}]{papineni2002bleu}
Papineni, K.; Roukos, S.; Ward, T.; and Zhu, W.-J.
\newblock 2002.
\newblock {BLEU}: a method for automatic evaluation of machine translation.
\newblock In {\em ACL}.

\bibitem[\protect\citeauthoryear{Popovi{\'c}}{2015}]{popovic2015chrf}
Popovi{\'c}, M.
\newblock 2015.
\newblock chrf: character n-gram f-score for automatic mt evaluation.
\newblock In {\em Proceedings of the Tenth Workshop on Statistical Machine
  Translation},  392--395.

\bibitem[\protect\citeauthoryear{Ranzato \bgroup et al\mbox.\egroup
  }{2016}]{Ranzato:2016:ICLR}
Ranzato, M.; Chopra, S.; Auli, M.; and Zaremba, W.
\newblock 2016.
\newblock Sequence level training with recurrent neural networks.
\newblock In {\em ICLR}.

\bibitem[\protect\citeauthoryear{Shaw, Uszkoreit, and
  Vaswani}{2018}]{shaw2018self}
Shaw, P.; Uszkoreit, J.; and Vaswani, A.
\newblock 2018.
\newblock {Self-Attention with Relative Position Representations}.
\newblock In {\em NAACL}.

\bibitem[\protect\citeauthoryear{Shen \bgroup et al\mbox.\egroup
  }{2016}]{Shen:2016:ACL}
Shen, S.; Cheng, Y.; He, Z.; He, W.; Wu, H.; Sun, M.; and Liu, Y.
\newblock 2016.
\newblock Minimum risk training for neural machine translation.
\newblock In {\em ACL}.

\bibitem[\protect\citeauthoryear{Shen \bgroup et al\mbox.\egroup
  }{2018}]{Shen:2018:AAAI}
Shen, T.; Zhou, T.; Long, G.; Jiang, J.; Pan, S.; and Zhang, C.
\newblock 2018.
\newblock {DiSAN: directional self-attention network for RNN/CNN-free language
  understanding}.
\newblock In {\em AAAI}.

\bibitem[\protect\citeauthoryear{Sutskever, Vinyals, and
  Le}{2014}]{sutskever2014sequence}
Sutskever, I.; Vinyals, O.; and Le, Q.~V.
\newblock 2014.
\newblock Sequence to sequence learning with neural networks.
\newblock In {\em NIPS}.

\bibitem[\protect\citeauthoryear{Tu \bgroup et al\mbox.\egroup
  }{2016}]{tu2016modeling}
Tu, Z.; Lu, Z.; Liu, Y.; Liu, X.; and Li, H.
\newblock 2016.
\newblock Modeling coverage for neural machine translation.
\newblock In {\em ACL}.

\bibitem[\protect\citeauthoryear{{Tu} \bgroup et al\mbox.\egroup
  }{2017}]{Tu:2017:AAAI}
{Tu}, Z.; {Liu}, Y.; {Shang}, L.; {Liu}, X.; and {Li}, H.
\newblock 2017.
\newblock Neural machine translation with reconstruction.
\newblock In {\em AAAI}.

\bibitem[\protect\citeauthoryear{Vaswani \bgroup et al\mbox.\egroup
  }{2017}]{vaswani2017attention}
Vaswani, A.; Shazeer, N.; Parmar, N.; Uszkoreit, J.; Jones, L.; Gomez, A.~N.;
  Kaiser, {\L}.; and Polosukhin, I.
\newblock 2017.
\newblock Attention is all you need.
\newblock In {\em NIPS}.

\bibitem[\protect\citeauthoryear{Williams}{1992}]{williams1992simple}
Williams, R.~J.
\newblock 1992.
\newblock Simple statistical gradient-following algorithms for connectionist
  reinforcement learning.
\newblock {\em Machine Learning} 8(3-4):229--256.

\bibitem[\protect\citeauthoryear{Wiseman and Rush}{2016}]{Wiseman:2016:EMNLP}
Wiseman, S., and Rush, A.~M.
\newblock 2016.
\newblock Sequence-to-sequence learning as beam-search optimization.
\newblock In {\em EMNLP}.

\bibitem[\protect\citeauthoryear{Wu \bgroup et al\mbox.\egroup
  }{2016}]{Wu:2016:arXiv}
Wu, Y.; Schuster, M.; Che, Z.; and Le, Q. V. e.~a.
\newblock 2016.
\newblock Google's neural machine translation system: Bridging the gap between
  human and machine translation.
\newblock {\em arXiv}.

\bibitem[\protect\citeauthoryear{Wu \bgroup et al\mbox.\egroup
  }{2017}]{wu2017adversarial}
Wu, L.; Xia, Y.; Zhao, L.; Tian, F.; Qin, T.; Lai, J.; and Liu, T.-Y.
\newblock 2017.
\newblock Adversarial neural machine translation.
\newblock {\em arXiv}.

\bibitem[\protect\citeauthoryear{Yang \bgroup et al\mbox.\egroup
  }{2018}]{Yang:2018:NAACL}
Yang, Z.; Chen, W.; Wang, F.; and Xu, B.
\newblock 2018.
\newblock Improving neural machine translation with conditional sequence
  generative adversarial nets.
\newblock In {\em NAACL}.

\bibitem[\protect\citeauthoryear{Yu \bgroup et al\mbox.\egroup
  }{2017}]{Yu:2017:AAAI}
Yu, L.; Zhang, W.; Wang, J.; and Yu, Y.
\newblock 2017.
\newblock {SeqGAN: sequence generative adversarial nets with policy gradient}.
\newblock In {\em AAAI}.

\bibitem[\protect\citeauthoryear{Zhang \bgroup et al\mbox.\egroup
  }{2017}]{zhang2017thumt}
Zhang, J.; Ding, Y.; Shen, S.; Cheng, Y.; Sun, M.; Luan, H.; and Liu, Y.
\newblock 2017.
\newblock {THUMT}: {An} open source toolkit for neural machine translation.
\newblock {\em arXiv preprint arXiv:1706.06415}.

\bibitem[\protect\citeauthoryear{Zheng \bgroup et al\mbox.\egroup
  }{2018}]{Zheng:2018:TACL}
Zheng, Z.; Zhou, H.; Huang, S.; Mou, L.; Xinyu, D.; Chen, J.; and Tu, Z.
\newblock 2018.
\newblock Modeling past and future for neural machine translation.
\newblock {\em TACL}.

\end{thebibliography}
\bibliographystyle{aaai}
\end{document}